\documentclass[sigconf]{acmart}

\AtBeginDocument{%
  \providecommand\BibTeX{{%
    \normalfont B\kern-0.5em{\scshape i\kern-0.25em b}\kern-0.8em\TeX}}}

\copyrightyear{2020}
\acmYear{2020}
\setcopyright{acmcopyright}\acmConference[CIKM '20]{Proceedings of the 29th ACM International Conference on Information and Knowledge Management}{October 19--23, 2020}{Virtual Event, Ireland}
\acmBooktitle{Proceedings of the 29th ACM International Conference on Information and Knowledge Management (CIKM '20), October 19--23, 2020, Virtual Event, Ireland}
\acmPrice{15.00}
\acmDOI{10.1145/3340531.3412104}
\acmISBN{978-1-4503-6859-9/20/10}

\usepackage{microtype}
\usepackage{url}
\usepackage{mathrsfs}
\usepackage[linesnumbered,ruled,vlined]{algorithm2e}
\usepackage{multirow}
\usepackage{graphicx} 
\usepackage{float} 
\usepackage{subfigure} 
\usepackage{color}
\usepackage{enumerate}
\usepackage{booktabs}
\usepackage[normalem]{ulem}
\usepackage[toc,page,title]{appendix}

\usepackage{xcolor}

\usepackage{xspace}
\newcommand{\cparagraph}[1]{\vspace{1mm}\noindent\textbf{#1}}

\begin{document}

\title{NASE: Learning Knowledge Graph Embedding for Link Prediction via Neural Architecture Search}

\author{Xiaoyu Kou}
\email{kouxiaoyu@pku.edu.cn}
\affiliation{%
  \institution{Peking University}
  \city{Beijing Shi}
  \country{China}
}

\author{Bingfeng Luo}
\email{luobingfeng981@pingan.com.cn}
\affiliation{%
  \institution{Pingan Life Insurance of China}
  \country{Shenzhen Shi, China}}

\author{Huang Hu}
\email{huahu@microsoft.com}
\affiliation{%
  \institution{Microsoft Corporation}
  \city{Beijing Shi, China}
}

\author{Yan Zhang}
\email{zhyzhy001@gmail.com}
\affiliation{%
  \institution{Peking University}
  \state{Beijing Shi}
  \country{China}}



\begin{abstract}
Link prediction is the task of predicting missing connections between entities in the knowledge graph (KG). 
While various forms of models are proposed for the link prediction task, most of them are designed based on a few known relation patterns in several well-known datasets.
Due to the diversity and complexity nature of the real-world KGs, it is inherently difficult to design a model that fits all datasets well.
To address this issue, previous work has tried to use Automated Machine Learning (AutoML) to search for the best model for a given dataset. However, their search space is limited only to bilinear model families.
In this paper, we propose a novel Neural Architecture Search (NAS) framework for the link prediction task. 
First, the embeddings of the input triplet are refined by the \textit{Representation Search Module}.
Then, the prediction score is searched within the \textit{Score Function Search Module}.
This framework entails a more general search space, which enables us to take advantage of several mainstream model families, and thus it can potentially achieve better performance. 
We relax the search space to be continuous so that the architecture can be optimized efficiently using gradient-based search strategies.
Experimental results on several benchmark datasets demonstrate the effectiveness of our method compared with several state-of-the-art approaches. 

\end{abstract}



\maketitle
\section{Introduction}

Knowledge Graph (KG) consists of facts in the form of triplet $(h, r, t)$, where the head $h$ and tail $t$ are entities while the relation $r$ refers to different types of edges between entities.
In recent years, KG has been successfully applied to many fields~\cite{bordes2014open, moon2019opendialkg}. 
However, most existing KGs are incomplete and noisy, which severely limits their application in practice~\cite{dong2014knowledge}. 
To tackle this issue, the link prediction task has been proposed to predict the existence of links between any two entities in a KG~\cite{bordes2013translating}, which quickly becomes a fundamental and challenging task in the KG field.

One strand of existing link prediction models operates in a reconstructive way. They reconstruct the embedding of the head (or tail) of a triplet $(h, r, t)$ using the corresponding relation and tail (or head) embeddings, and calculate the plausibility of the triplet by measuring the difference between the original and the reconstructed embeddings. 
These works either model this relationship in an explainable way (e.g., TransE~\cite{bordes2013translating}, RotatE~\cite{sun2019rotate}), or utilize the black-box but expressive convolution operations (e.g., ConvE~\cite{dettmers2018convolutional}).
Another strand of works considers link prediction as a semantic matching problem~\cite{ji2020survey}. 
They take the embeddings of the head, relation and tail as input, and output a matching score for the elements in each triplet using bi-linear transformation (e.g., DistMult~\cite{yang2014embedding}, SimplE~\cite{kazemi2018simple}), convolution (e.g., ConvKB~\cite{nguyen2017novel}) and etc.

These works vary a lot in situations that they are suitable for, like the relation types in the KG, the sparsity of the KG, etc.
Therefore, choosing a suitable architecture for a specific KG often requires careful analysis of both the dataset and the model.
To tackle this issue, \citet{zhang2019autosf} proposes to use AutoML to greedily search for optimal score functions for distinct KGs.
However, their work is constrained to bilinear semantic matching models and does not include the reconstruction-based models into their search space.

In this paper, we propose a novel Neural Architecture Search (NAS) framework to search for the most effective architecture for a given dataset.
The framework entails a more general search space that contains both semantic matching models and reconstructive models.
Therefore, it has the potential to combine the strength of the two model families. 
Instead of searching over a discrete set of candidate architectures, we relax the search space to be continuous, so that the architecture can be optimized using the efficient gradient-based search algorithm.

\begin{figure}[htbp]
    \centering
    \vspace{-3mm}
    \includegraphics[width=0.8\linewidth]{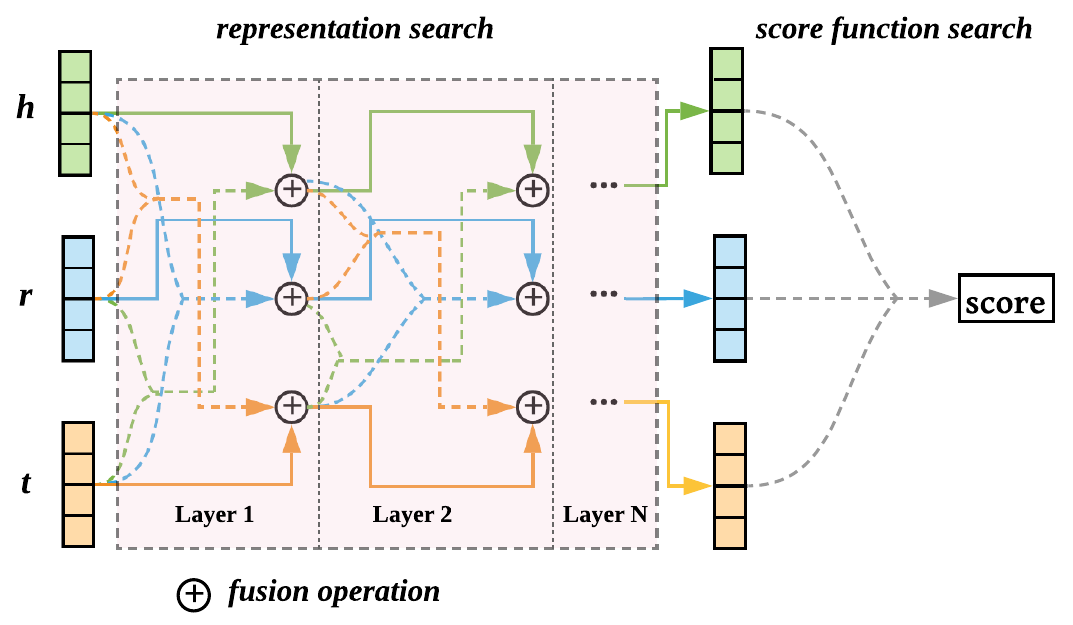}
    \vspace{-4mm}
    \caption{Overview of NASE. Solid lines indicate direct connections, while dashed lines indicate searchable connections that are determined by the architecture search process.}
    \label{fig:model_overview}
    \vspace{-5mm}
\end{figure}

As shown in Fig.~\ref{fig:model_overview}, our NAS framework contains two search modules.
The \textit{representation search module} aims to refine the embeddings $\boldsymbol{e}_h, \boldsymbol{e}_r, \boldsymbol{e}_t$ of the head, relation, and tail respectively through multiple representation layers.
The \textit{score function search module} is responsible for selecting a shallow architecture to calculate a plausibility score for the input triplet.
While the operators in each module have a broad range of choices, in this work, we primarily focus on a proof-of-concept of this two-level search space by constraining the operators to architectures that are representatives of existing link prediction models.
Specifically, we constrain the search space of the \textit{representation search module} to be reconstructive models, whose input and output have good compatibility of this module.
As for the \textit{score function search module}, we select representative models from mainstream model families in the link prediction task.
To avoid overfitting, we also add the identity operation in the \textit{representation search module} so that the NAS algorithm could choose to use the original $\boldsymbol{e}_h, \boldsymbol{e}_r, \boldsymbol{e}_t$, or even degenerate to the basic models in the \textit{score function search module} when necessary.

On the one hand, we can consider the representation search module refines $\boldsymbol{e}_h, \boldsymbol{e}_r, \boldsymbol{e}_t$ in a black-box way.
On the other hand, the output of the \textit{representation search module} may also embed the constraints modeled in the reconstruction-based models.
Therefore, the final score could potentially benefit from the cross-validation of multiple models, which will likely lead to better prediction results.

We evaluate our approach on several popular benchmark datasets.
Extensive experiments demonstrate that our approach has good generalization ability over different datasets, and achieves better performance than strong baseline models in most of the datasets.

\section{Approach}
\subsection{Problem Formulation}
Let $\mathcal{E}$, $\mathcal{R}$ denote the entity set and the relation set in a knowledge graph $G$ which can be formalised as a set of triplets $\{(h, r, t)\}\subseteq \mathcal{E} \times \mathcal{R} \times \mathcal{E}$. 
Given a triplet $(h, r, t)$, we denote the embeddings of them as $\boldsymbol{e_h}$, $\boldsymbol{e_r}$, $\boldsymbol{e_t}\in \mathbb{R}^{d}$. 
The link prediction task can be formulated as maximizing a function $f({h}, {r}, {t})$ over all triplets, which is expected to give higher scores for valid triplets than invalid ones.
Our work aims to find the best performing architecture of $f({h}, {r}, {t})$ given a pre-defined architecture space via Network Architecture Search (NAS) for a specific dataset. 

\vspace{-0.5em}
\subsection{Search Space for Link Prediction} \label{sec:search_space}
We name our framework as NASE, which refers to NAS-based knolwedge graph embedding for link prediction.
As illustrated in Fig.~\ref{fig:model_overview}, NASE utilizes two search modules.
The \textit{representation search module} aims to generate more informative embeddings for head, relation, and tail, while the goal of the \textit{score function search module} is to find a suitable architecture to integrate these embeddings to produce the final plausibility score for the $(h, r, t)$ triplet.
The representation part searches among several reconstructive models. And the search space of the score function is more flexible that it contains both reconstruction based and semantic matching based ones.
Since the resulting architecture may contain multiple base models, our framework potentially has the ability to combine the strength of several model families. 

\subsubsection{Representation Search Module}
In the representation search module, we include the convolution operators and translation operators in our search space, which correspond to two mainstream model families of the reconstruction-based link prediction models.

As shown in Fig.~\ref{fig:model_overview}, each representation layer takes the embeddings of the head, relation, and tail as input, and produce a set of new embeddings. 
Each operator is followed by a fusion step that linearly combines the new embeddings of the head, relation, or tail with the original ones. For example, we define the fusion step of the head embedding as follow:
\begin{align}
\small
    \boldsymbol{e}_h^{l+1} &= \beta_h^l \boldsymbol{e_h^l} + (1 - \beta_h^l) \mathrm{opt}(\boldsymbol{e}_r^l , \boldsymbol{e}_t^l), \\
    \beta_h^l &= \sigma(\boldsymbol{W}_h^l [\boldsymbol{e}_h^l;\mathrm{opt}(\boldsymbol{e}_r^l , \boldsymbol{e}_t^l)] + b_h^l),
\end{align}
where $\boldsymbol{e}_h^l$ is the head embedding output at layer $l$, $\mathrm{opt}(\cdot,\cdot)$ denotes an searchable operator, $\boldsymbol{W}_h^l$ and $b_h^l$ are trainable parameters. 
The goal of introducing the fusion step is to force the new embedding to focus on the information about the head, rather than things that should be modeled by relation or tail embeddings.  
Likewise, this transformation is also applied to $\boldsymbol{e}_r^l$ and $\boldsymbol{e}_t^l$.
Specifically, we enumerate the three categories of candidate operators as follows:

\cparagraph{Convolution Operators.} 
Similar to ConvE~\cite{dettmers2018convolutional}, we use 1D convolution with filter sizes 2, 4, and 2D convolution with filter sizes 3, 5 as candidate operators. 
\begin{align}
\small
    \boldsymbol{e}_h^{l+1} = \text{ReLU}\big(\text{Conv1d} ([\boldsymbol{e}_r^{l} ; \boldsymbol{e}_t^{l}]) \big),\   \boldsymbol{e}_h^{l+1} = \text{ReLU}\big(\text{Conv2d} ([\overline{\boldsymbol{e}_r^{l}} ;\ \overline{\boldsymbol{e}_t^{l}}]) \big),
\end{align}
where $[\cdot ;\cdot]$ is row-wise concatenation, $\overline{\boldsymbol{e}}$ denote 2D reshaping of $e$.

\cparagraph{Translation Operators.} 
Translation-based models follow the restriction of:
\begin{equation} 
\vspace{-1mm}
g_{r,1}(\boldsymbol{e_h}) + \boldsymbol{e_r} - g_{r,2}(\boldsymbol{e_t}) = \boldsymbol{0}
\label{eq:translation-constraint}
\end{equation}
where $g_{r,\cdot}(x)$ is a model-specific transformation function.
$\boldsymbol{e_h}, \boldsymbol{e_r}, \boldsymbol{e_t}$ can be inferred using Eq.~\ref{eq:translation-constraint} given the other two embeddings and the inverse function of $g_{r,\cdot}(x)$.
In this work, we use $g_{r,\cdot}(x)=\textbf{W}\textbf{x}$, and $\textbf{W}$ can be either an identity matrix or an unconstrained one.
Therefore, our search space includes two famous translation-based models: TransE~\cite{bordes2013translating} and TransR~\cite{lin2015learning}.

\cparagraph{Identity Operators.}
The identity operator directly maps the input to output without any transformations. 
The purpose of introducing such an operator is to prevent the resulting architecture from being too complicated, and it enables the final model to degenerate to basic models in the \textit{score function search space.} 

\subsubsection{Score Function Search Module}
In this module, we search among several semantic matching based models to produce a final plausibility score, which is used to predict whether the triplet is valid.
While we can include as many models as we want, we only include 1-2 representative models in each model family for proof-of-concept.
The experimental results show that this search space already produces satisfying results.

\cparagraph{Convolution-Based Score Function.}
Our convolution-based score function has a similar form to ConvKB~\cite{nguyen2017novel}:
\begin{align}
    f(h, r, t) = \boldsymbol{W} \left( \text{ReLU}\big(\text{Conv} ([\boldsymbol{e_h} ; \boldsymbol{e_r} ; \boldsymbol{e_t}]) \big) \right),
\end{align}
where $\text{Conv}(\cdot)$ indicates the convolution layer with $M$ $3 * 3$ filters and $\boldsymbol{W} \in \mathbb{R}^{1\times {Md}}$ is a trainable matrix.

\cparagraph{Translation-Based Score Function.}
We use the classic TransE~\cite{bordes2013translating} as the translation-based score function: 
\begin{align}
    \vspace{-2mm}
    f(h, r, t) = ||\boldsymbol{e_h} + \boldsymbol{e_r} - \boldsymbol{e_t}||_p,
\end{align}
where $||\cdot||_p$ denotes the $p$-norm.

\cparagraph{Bilinear Score Function.}
We adopt two bilinear models: DistMult~\cite{yang2014embedding} (Eq.~\ref{distmult}), and SimplE~\cite{kazemi2018simple} (Eq.~\ref{SimplE}):
\begin{equation} 
\label{distmult}
\vspace{-2mm}
    f(h, r, t) = \boldsymbol{e}_h^T \boldsymbol{M_r} \boldsymbol{e_t},
\end{equation}
\begin{equation}    
\label{SimplE}
    f(h, r, t) = 1/2 \left( \boldsymbol{e}_h^T \boldsymbol{M}_r, \boldsymbol{e}_t + \boldsymbol{e}_h^T, \boldsymbol{M}_r', \boldsymbol{e}_t \right),
\end{equation}
where $\boldsymbol{M_r}$ is the diagonal matrix generated from $\boldsymbol{e_r}$, and $\boldsymbol{M_r'}$ is another randomly initialzed diagonal matrix for relation $r$.

\cparagraph{MLP Score Function.}
Each triplet representation $(\boldsymbol{e_h}, \boldsymbol{e_r}, \boldsymbol{e_t})$ is concatenated into a single feature vector.
The feature vector is then fed into a single-hidden-layer MLP to calculate the final score.

\subsection{Search Procedure of NASE}
\label{sec:search_proc}
We use the gradient-based search strategy proposed by DARTS~\cite{liu2018darts} as the architecture search algorithm.
Specifically, if a function $g = \varphi(x)$ can be instantiated by $k$ candidate operations $\varphi_1, \varphi_2, ..., \varphi_k$, we then define $k$ corresponding operation weights $\alpha_1, \alpha_2, ..., \alpha_k$, and relaxes the original equation as follows in the search phase:
\begin{equation}
\small
g = \sum_{i=1}^{k}{a_i \varphi_i(x)}, \quad\quad a_i = \frac{\mathrm{exp}(\alpha_i)}{\sum_{i=1}^{k}{\mathrm{exp}(\alpha_i)}}
\end{equation}
In NASE, as shown in the dashed hyperedges that links inputs and outputs in Fig.~\ref{fig:model_overview}, we need to search for the functions used to calculate $\boldsymbol{e_h}, \boldsymbol{e_r}, \boldsymbol{e_t}$ in each representation layer, and the final score function.  
The candidates for each dashed hyperedge have their own operation weights.

The model weights $\boldsymbol{\theta}$ and the operation weights $\boldsymbol{\alpha}$ are updated iteratively.
For each batch, we first update $\boldsymbol{\theta}$ with $\boldsymbol{\alpha}$ fixed using SGD.
Then, $\boldsymbol{\alpha}$ is updated using the new $\boldsymbol{\theta}$.
Both $\boldsymbol{\theta}$ and $\boldsymbol{\alpha}$ are updated according to the following loss function:
\begin{align}
    \small
    \mathcal{L}=-\frac{1}{N}\sum_{i=1}^{N}(y_i \log(f(\cdot)) + (1-y_i)(1- \log(f(\cdot))),
\end{align}
where the label $y_i$ is 1 if the triplet $(h,r,t)$ is valid and 0 otherwise, $f(\cdot)$ is the architecture that produces the plausibility score.
After convergence, the candidates with the most significant weights are selected to form the final architecture.
Finally, the resulting architecture is then trained from scratch.

\section{Experiments}

\subsection{Experimental Setup}

\cparagraph{Datasets}
We evaluate our model on five public benchmark datasets: FB15k-237~\cite{toutanova2015observed}, WN18RR~\cite{bordes2013translating}, Medical-E~\footnote{We construct Medical-E by extracting drug and disease entities from the Freebase with five relations: treatments, symptoms, risk factors, causes, and prevention factors.}, Medical-C~\footnote{https://github.com/liuhuanyong/QASystemOnMedicalKG}, and Military~\footnote{http://openkg.cn/dataset/techkg10}.
Existing KG embedding methods are usually designed based on the commonly used FB15k-237 and WN18RR datasets.
Therefore, we include the other three less common datasets to better compare the dataset adaptation ability of each system.
The statistics of these KGs are shown in Table~\ref{tab:Statistics}. 

\begin{table}[htbp]
    \renewcommand\arraystretch{0.9}
    \vspace{-3mm}
    \centering
    \begin{tabular}{rrrrrr}
    \hline
     Dataset & Entity & Relation & Train & Validation & Test  \\ \hline
     FB15k-237 & 14,541 & 237 & 272,115 & 17,535 & 20,466  \\
     WN18RR & 40,943 & 11 & 86,835 & 3,034 & 3,134 \\
     Medical-E & 1,516 & 5 & 5,000 & 1,000 & 1,000 \\
     Medical-C & 22,119 & 6 & 156,367 & 13,682 & 24,048 \\
     Military & 11,975 & 15 & 92,016 & 11,502 & 11,502  \\
     \hline
    \end{tabular}
    \vspace{1mm}
    \caption{Statistics of Each Dataset. }
    \label{tab:Statistics}
    \vspace{-9.5mm}
\end{table}

\cparagraph{Training Details}
We perform grid search for hyper-parameters: number of representation layers $N=$ \{1, 2, 3, 4\}, embedding dimension $d=$ \{100, 200, 400\}, learning rate $lr=$ \{1e-2, 1e-3, 1e-4\}, batch size $M=$ \{128, 256\}. 
We find that the following combination performs best in most datasets: $N=1$, $d=400$, $lr=1e-3$, $M=128$.

\cparagraph{Baselines}
We compare NASE with several state-of-the-art models:
(1) Reconstruction-Based Models: TransE~\cite{bordes2013translating}, TransR~\cite{lin2015learning}, ConvE~\cite{dettmers2018convolutional} and RotatE~\cite{sun2019rotate}.
(2) Semantic Matching Models: DistMult~\cite{yang2014embedding}, ComplEx~\cite{trouillon2016complex}, SimplE~\cite{kazemi2018simple}, TuckER~\cite{balavzevic2019tucker} and ConvKB~\cite{nguyen2017novel}.
(3) AutoML System: AutoSF~\cite{zhang2019autosf}.
Results are taken from published papers, if possible. 
Other results are produced by running publicly released codes, and we tune the hyperparameters via grid search.

\subsection{Results}
\begin{table*}[]
    \renewcommand\arraystretch{0.9}
    \centering
    \begin{tabular}{c|ccc|ccc|ccc|ccc|ccc}
    \hline
         \multirow{2}{*}{~} & \multicolumn{3}{c|}{\textbf{FB15k-237}} & \multicolumn{3}{c}{\textbf{WN18RR}} & \multicolumn{3}{|c|}{\textbf{Medical-E}} &
         \multicolumn{3}{c|}{\textbf{Medical-C}} &
         \multicolumn{3}{c}{\textbf{Military}} \\ 
         \cline{2-16}
                    & MR    & MRR   & H@10   & MR    & MRR   & H@10 & MR    & MRR   & H@10 & MR    & MRR   & H@10 & MR    & MRR   & H@10   \\ \hline
         TransE     & 357   & .294  & .465      & 3384  & .226  & .501  & 289 & .350 & .425 & 1061 & .304 & .391 & 198 & .387 & .618 \\
         TransR     & 349   & .301  & .461      & 3317  & .219  & .498  & 301 & .352 & .421 & 1059 & .302 & .401 & 192 & .381 & .599   \\
         RotatE     & \underline{177}   & .338  & .533      & 3340 & \underline{.476}  & \textbf{.571} & \underline{149} & .437 & .541 & 1122 & .320 & .412 & \underline{155} & .419 & .633 \\ 
         ConvE      & 244   & .325  & .501      & 5277 & .460  & .480 & 247 & .399 & .458 & 1739 & .321 & .397 & 305 & .391 & .561 \\  \hline
         DistMult   & 254   & .241  & .419      & 5110  & .430  & .490 & 271 & .362 & 429 & 1273 & .346 & .427 & 187 & .380 & .592  \\
         ComplEx    & 339   & .247  & .428      & 5261 & .440  & .510 & 207 & .367 & .496 & 1577 & .335 & .419 & 487 & .367 & .496 \\
         SimplE     & 203  & .341   & 534       & 3561  & .462   & .550 & 154 & .429 & .539 & 1108 & .313 & .440 & 167 & .402 & \underline{.635}   \\ 
         TuckER     & -    & .358   &.544       & -     & .470  & .526 & 152 & .445 & \underline{.545} & 1010 & .350 & .423 & 161 & .399 & .612  \\ 
         ConvKB     & 257   & \underline{.396}  & .517      & \underline{2554} & .248  & .525 & 150 & .361 & .521 & 989 & .341 & \underline{.449} & 160 & \underline{.420} & .624\\  \hline
         AutoSF     & -     & .360  & \underline{.552}      & -     & \textbf{.490} & \underline{.567} & 151 & \underline{.451} & .537 & \underline{896} & \underline{.352} & .440 & 231 & .418 & .606    \\ \hline
         NASE       & \textbf{170} & \textbf{.421} & \textbf{.575} & \textbf{2147} & .465 & .553 & \textbf{147} & \textbf{.525} & \textbf{.588} & \textbf{765} & \textbf{.360} & \textbf{.461} & \textbf{154} & \textbf{.440} & \textbf{.654} \\
    \hline
    \end{tabular}
    \vspace{1.5mm}
    \caption{Link prediction results. Bold and underlined numbers refer to the best and second-best results respectively.}
    \label{tab:result}
    \vspace{-8mm}
\end{table*}

As can be seen in Table~\ref{tab:result}, NASE clearly outperforms AutoSF in four out of five datasets, which demonstrates the effectiveness of our architecture.
However, we also observe that our model is inferior to AutoSF in the WN18RR dataset except for the MR metric.
The main reason for this phenomenon is that, as shown in Table~\ref{tab:Statistics}, the WN18RR dataset is the sparsest one in the five datasets.
Besides, we can also see that there is no absolute winner among human-designed models.
On the other side, NASE achieves better or comparable results to them except for RotatE in WN18RR.
Note that FB15k-237 and WN18RR are two of the most commonly used datasets in link prediction, which means their characteristics are well studied by existing link prediction works. 
Therefore, achieving the best results in FB15k-237 and all other less common datasets confirms that NASE has a good generalization ability to different datasets with various characteristics.
We do not include RotatE in our search space because it requires $\textbf{e}_r$ to have different dimension from $\textbf{e}_h$ and $\textbf{e}_t$. We leave this to the future work.

\begin{figure}[!t]
    \centering
    \subfigure[Military]{
        \begin{minipage}[t]{0.4\linewidth}
        \includegraphics[width=3.5cm]{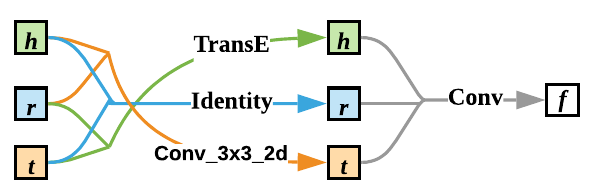}
        \vspace{-1em}
        \label{fig:arc_1}
        \end{minipage}%
    }%
    \subfigure[Medical-E]{
        \begin{minipage}[t]{0.6\linewidth}
        \includegraphics[width=5.0cm]{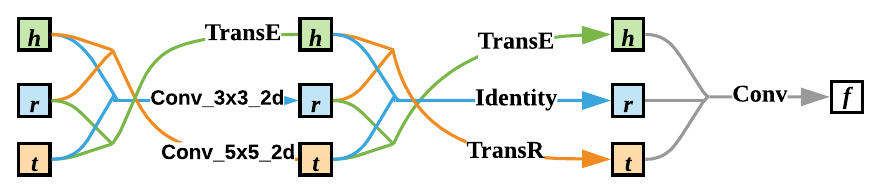}
        \vspace{-1em}
        \label{fig:arc_2}
        \end{minipage}%
    }%
    \vspace{-2em}
    \caption{Two architectures searched on by NASE.
    }
    \vspace{-1.5em}
    \label{fig:arc}
\end{figure} 

Fig.~\ref{fig:arc} shows two architectures searched by NASE.
We observe that most of the resulting architectures only utilize one representation layer, which is consistent with the observations in previous works that deep architectures will usually cause overfitting problem.
The only exception occurs in the Medical-E dataset, where the resulting architecture contains two representation layers. 
We conjecture that this is due to the simplicity of the Medical-E dataset.
It only has two types of entities with five relation types.
The test set and train set are also identically distributed.
Therefore, overfitting is a less severe problem than underfitting in this dataset.

\subsection{Ablation Study}
\begin{table}[ht]
    \vspace{-1em}
    \centering
    \renewcommand\arraystretch{0.78}
    \begin{tabular}{lccc}
    \toprule
         Ablation & MR & MRR & H@10   \\ \midrule
         Full NASE & \textbf{170} & \textbf{.421} & \textbf{.575} \\  \midrule
         (-) Representation search module & 201 & .395 & .566 \\
         (-) Score function search module & 229 & .389 & .559 \\
         (-) Fusion step & 251 & .380 & .550 \\
         \bottomrule
    \end{tabular}
    \vspace{0.5em}
    \caption{Ablation study on FB15k-237 dataset by NASE. }
    \label{tab:Ablation}
    \vspace{-2em}
\end{table}

To demonstrate the influence of the components in our NAS framework, we also conduct an ablation study using FB15k-237.
As shown in Table~\ref{tab:Ablation}, removing the representation search module clearly decreases the performance in all metrics, which proves the effectiveness of our representation search module.
Besides, 
as shown in the third row of Table~\ref{tab:Ablation}, removing the score function search module also decreases the performance, but it still performs much better than vanilla TransE (see Table~\ref{tab:result}). 
This again indicates that the representation search module can produce more informative representations than the original one.
Moreover, we also examine the performance of the fusion step by replacing the linear combination with direct addition.
The drop indicates that using learnable weights to linearly combine the input and output vectors in the representation layers plays an essential role in learning high-quality representations.

\section{Conclusion and Future Work}
In this paper, we propose a novel NAS framework for the link prediction task, which can combine the strength of both reconstruction based and semantic matching based models. 
Experimental results show that NASE outperforms several state-of-the-art human-designed models and AutoML based models in most of the datasets.
In future work, we would like to explore the possibility of more general search spaces to include more strong architectures.

\section{Acknowledgments}
This work is supported by National Key Research and Development Program of China under Grant No. 2018AAA0101902, NSFC under Grant No. 61532001, and MOE-ChinaMobile Program under Grant No. MCM20170503.

\bibliographystyle{ACM-Reference-Format}
\bibliography{sample-base}

\appendix

\settopmatter{printacmref=true}
\end{document}